\documentclass[num-refs]{wiley-article}

\usepackage{siunitx}
\newtheorem{myDef}{Definition}
\newtheorem{mypro}{Problem}
\usepackage{subfigure} 
\usepackage{tabularx}
\usepackage{array}
\usepackage{boldline, multirow}
\usepackage{float}

\usepackage{algorithm}
\usepackage{algorithmic}

\definecolor{YUMColor}{rgb}{0.9, 0.1, 0.0}

\definecolor{BlueColor}{rgb}{0.15, 0.3, 1.0}
\newcommand{\BL}[1]{{\color{BlueColor} #1}}

\papertype{Research Article}

\title{Deep Dual Support Vector Data Description for Anomaly Detection on Attributed Networks}

\author[1]{Fengbin Zhang}
\author[2\authfn{1}]{Haoyi Fan}
\author[1]{Ruidong Wang}
\author[3]{Zuoyong Li}
\author[4\authfn{1}]{Tiancai Liang}

\affil[1]{School of Computer Science and Technology, Harbin University of Science and Technology, Harbin, China.}
\affil[2]{School of Information Engineering, Zhengzhou University, Zhengzhou, China.}
\affil[3]{Fujian Provincial Key Laboratory of Information Processing and Intelligent Control, Minjiang University, Fuzhou, China.}
\affil[4]{GRG Banking Technology Co., Ltd, Guangzhou, China.}

\corraddress{\authfn{1}Haoyi Fan, School of Information Engineering, Zhengzhou University, Zhengzhou 450001, China. \\
Email: isfanhy@gmail.com \\
\authfn{1}Tiancai Liang, GRG Banking Technology Co., Ltd, Guangzhou 510430, China. }
\corremail{13808895230@139.com}

\fundinginfo{National Natural Science Foundation of China, Grant/Award Number: 61972187 and 61172168;
Natural Science Foundation of Fujian Province, Grant/Award Number: 2020J02024 and 2019J01756.
Fuzhou Science and Technology Project: 2020-RC-186.
}

\runningauthor{Zhang et al.}

\begin{document}

\begin{frontmatter}
\maketitle

\begin{abstract}
Networks are ubiquitous in the real world such as social networks and communication networks, and anomaly detection on networks aims at finding nodes whose structural or attributed patterns deviate significantly from the majority of reference nodes. However, most of the traditional anomaly detection methods neglect the relation structure information among data points and therefore cannot effectively generalize to the graph structure data. In this paper, we propose an end-to-end model of Deep Dual Support Vector Data description based Autoencoder (Dual-SVDAE) for anomaly detection on attributed networks, which considers both the structure and attribute for attributed networks. Specifically, Dual-SVDAE consists of a structure autoencoder and an attribute autoencoder to learn the latent representation of the node in the structure space and attribute space respectively. Then, a dual-hypersphere learning mechanism is imposed on them to learn two hyperspheres of normal nodes from the structure and attribute perspectives respectively. Moreover, to achieve joint learning between the structure and attribute of the network, we fuse the structure embedding and attribute embedding as the final input of the feature decoder to generate the node attribute. Finally, abnormal nodes can be detected by measuring the distance of nodes to the learned center of each hypersphere in the latent structure space and attribute space respectively. Extensive experiments on the real-world attributed networks show that Dual-SVDAE consistently outperforms the state-of-the-arts, which demonstrates the effectiveness of the proposed method. 

\end{abstract}
\end{frontmatter}

\section{Introduction}

Graph structure data exists widely in the real world such as social media network \cite{liao2018attributed}, traffic network \cite{cui2019traffic} and product co-purchase network \cite{shi2018heterogeneous,fan2021heterogeneous}. However, real-world networks are usually noisy, and existing anomalous nodes will disrupt the performance of the conventional machine learning algorithms for different downstream data analysis tasks. For example, in a citation network, there are some papers that cite a few eccentric references which do not comply with the research content of the paper \cite{bandyopadhyay2019outlier}. In a mobile cashless payment platform, newly registered malicious accounts on some devices tend to be active only in a short term \cite{liu2018heterogeneous}.

\begin{figure}[bt]
\centering
\includegraphics[width=5in]{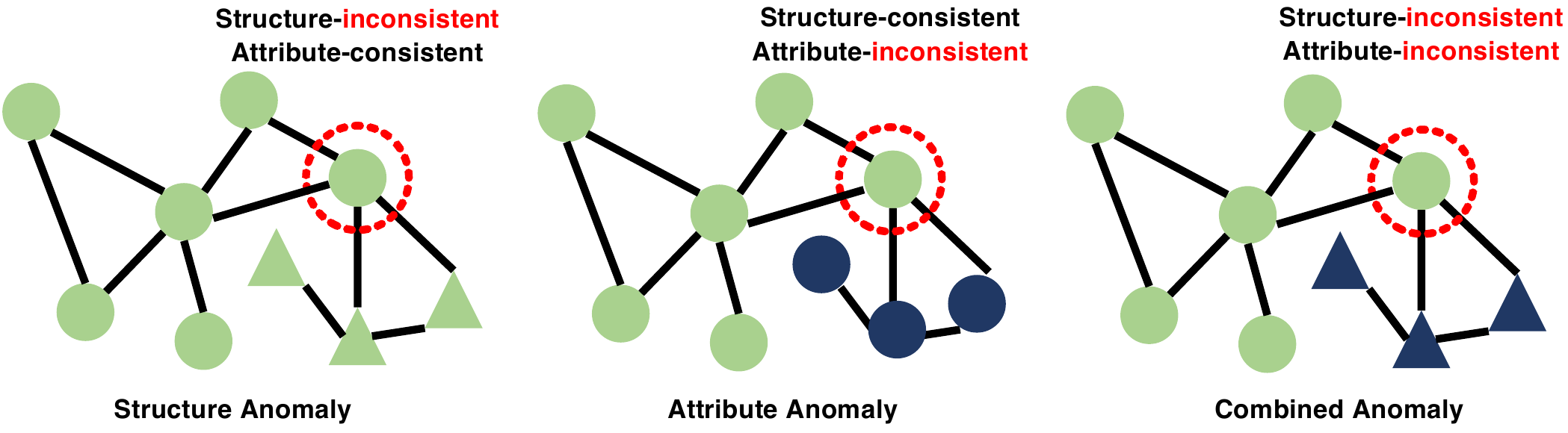}
\caption{Three types of anomalies in the attributed networks. The same shape of node indicates that the nodes belong to the same community, while the same color indicates that the nodes are consistent in terms of attribute.}
\label{fig:anomaly_types}
\end{figure}

Anomaly detection on attributed networks is a task to identify the nodes whose behaviors significantly differ from the other nodes, which has a broad impact on various domains such as network intrusion detection \cite{ding2012intrusion}, system fault diagnosis \cite{cheng2016ranking}, and social spammer detection \cite{fakhraei2015collective}. Normal nodes are usually consistent with their neighbors in the behavior from both link structure and node attribute perspective. Differently, anomalous nodes are usually categorized into three types \cite{bandyopadhyay2019outlier}: structure anomaly, attribute anomaly, and combined anomaly. As shown in Fig. \ref{fig:anomaly_types}, the nodes with structure anomaly are usually attribute-consistent but structure-inconsistent in the local community, while the attribute anomaly nodes are usually structure-consistent but attribute-inconsistent in the local community, and there are also some combined anomalies that are inconsistent in terms of both structure and attribute.

Recently, there is a growing research interest in anomaly detection on attributed networks. Some of them conduct anomaly detection using only structure information in the community-level by comparing the current node with other reference nodes within the same community \cite{perozzi2016scalable} or measuring the quality of connected subgraphs \cite{perozzi2018discovering}. Some of them study the problem of feature-level anomalies detection through subspace selection of node feature \cite{sanchez2013subspaces, perozzi2014focused}. While some residual analysis based methods \cite{li2017radar, peng2018anomalous} and graph autoencoder based methods \cite{ding2019deep, li2019specae, fan2020anomalydae} consider both the context and feature information to discover anomalies by assuming that anomalies cannot be approximated from other reference nodes, they use residual estimation or network reconstruction to measure the abnormality of each node. Although those exiting method had their fair share of success, there are still some shortcomings among them. Some of them rely on shallow learning mechanisms which are difficult to process the complex interactions between structure information and attribute information, especially high-dimensional complex data. While those deep-learning based methods are limited by the autoencoder architecture. Those methods aim at minimizing the reconstruction errors for the whole network, which will be impacted by the noisy node and have a potential problem of overfitting for both normal nodes and abnormal nodes.

To alleviate above-mentioned problems, inspired by the successful application of hypersphere learning in anomaly detection \cite{scholkopf2001estimating, tax2004support, wang2021one}, in this paper, we propose an end-to-end one-class classification based anomaly detection framework for attributed networks, named \textbf{dual} \textbf{s}upport \textbf{v}ector \textbf{d}ata description based \textbf{a}uto\textbf{e}ncoder (\textbf{Dual-SVDAE}), which aims at learning the compact hypersphere boundary of normal nodes' latent space from both the structure and attribute perspectives. Specifically, Dual-SVDAE consists of a structure autoencoder and an attribute autoencoder to learn the latent representation of the node in the structure space and attribute space respectively. Then,  a dual-hypersphere learning mechanism is imposed on them to learn two hyperspheres of normal nodes from the structure and attribute perspectives respectively. Moreover, to achieve joint learning between the structure and attribute among nodes, we fuse the structure embedding and attribute embedding as the final input of the feature decoder to generate the node attribute. Finally, abnormal nodes are detected by measuring the distance of nodes to the learned center of each hypersphere in the latent structure space and attribute space respectively. Extensive experiments on multiple real-world attributed networks show that Dual-SVDAE consistently outperforms the state-of-the-arts, which demonstrates the effectiveness of the proposed method.

The main contributions of this paper are outlined as follows:
\begin{itemize}
    \item We propose an end-to-end one-class classification based anomaly detection framework, named dual support vector data description based autoencoder (Dual-SVDAE), which is capable of learning the compact hypersphere boundary of normal nodes' latent space from both the structure and attribute perspectives for more robust anomaly detection on attributed networks.
    \item We design a joint learning mechanism to capture the interaction between structure and attribute during the learning of the corresponding hyperspheres, in which, a fusion module is designed to fuse the learned structure embedding and attribute embedding for node attribute generation. 
    
    \item We comprehensively evaluate the effectiveness of Dual-SVDAE on multiple real-world datasets, and the results demonstrate that our proposed method outperforms the state-of-the-arts. All source code and data{\footnote{{\url{https://haoyfan.github.io/}}}} will be released upon the acceptance of this paper.
\end{itemize}

\section{Related work}
In this section, we review the related works about the traditional anomaly detection and the attributed network based anomaly detection.

\subsection{Traditional Anomaly Detection}

Recently, most of the traditional anomaly detection methods are unsupervised which aims to detect anomalies without labeled data for the scenario that only a small number of labeled anomalous data combined with plenty of unlabeled data are available \cite{fan2020correlation}. Existing traditional anomaly methods can be categorized as: (a) clustering based methods, (b) reconstruction based methods, and (c) one-class classification based methods. Clustering based methods \cite{xiong2011group, zhang2018binary,zhang2018highly} estimate the density of data points and usually adopt a two-step strategy \cite{Varun2009survey} that conduct dimensionality reduction firstly and then clustering the normal data for anomaly detection. Reconstruction based methods assume that anomalies cannot be effectively reconstructed from the compressed low-dimensional projections, including PCA \cite{jolliffe2003principal} based approaches \cite{xu2010robust, pascoal2012robust} and autoencoder based approaches \cite{zhai2016deep, zhou2017anomaly, zong2018deep} that use reconstruction errors as the anomaly scores for detection. Different from the previously mentioned two categories, one-class classification based methods \cite{scholkopf2001estimating, tax2004support, perdisci2006using, amer2013enhancing, ruff2018deep} aims to learn a discriminative boundary between the normal and abnormal instances for anomaly detection. Among these methods, One-class SVM \cite{scholkopf2001estimating} tries to fit a tight hyperplane in the non-linearly transformed feature space to include more normal data. Inspired by One-class SVM, Support vector data description (SVDD) \cite{tax2004support} learns a hypersphere instead of a hyperplane around the normal data for more robust anomaly detection. Recently, some deep learning based one-class classifiers try to learn both the data representation and the hypersphere for anomaly detection. A typical method is Deep-SVDD \cite{ruff2018deep} which employs a neural network to extract features of data and construct a single classification objective function for anomaly detection. Besides, OCGAN \cite{perera2019ocgan} is an adversarial network based model for one-class novelty detection, which forces the latent space to have bounded support in the encoder and generate potential out-of-class examples for training. Based on OCGAN, OGN \cite{zaheer2020old} redefines the discriminator from identifying real and fake data to distinguishing the data reconstructions quality, which makes the discriminator detect subtle distortions of outliers' reconstruction. Although those methods achieve successful applications in the traditional anomaly detection domain, they can not generalize well to the graph data, where the structure patterns among data points are important \cite{wang2021one,zhang2020inductive}. Therefore, the problem of anomaly detection on graph data still be an open problem.

\subsection{Attributed Networks embedding}

Attributed networks are ubiquitous in the real world such as social networks and traffic networks. Recently, a lot of methods  \cite{huang2017label,cui2018survey} have been proposed which learn the latent node representation by using topological structure information and auxiliary information such as node or edge attribute information. Existing Attributed Networks embedding methods can be categorized as: (a) random walk \cite{perozzi2014deepwalk} based methods, (b) adjacency matrix decomposition based methods, and (c) deep learning based methods. Among the random walk based methods, SANE \cite{wang2018united} learns the low dimensional representations by using random walk to generate node sequences and using attention mechanisms to aggregate the information of neighbor nodes. TADW  \cite{yang2015network} incorporates the DeepWalk and text features into an integrated framework for network representation learning. Based on TADW, HSCA  \cite{zhang2016homophily} adds a regularization of neighbor nodes to capture the homogeneity of network. Besides, some methods based on matrix decomposition such as AANE \cite{huang2017accelerated} and BANE \cite{yang2018binarized} capture the data correlation between the structure and attribute by designing a proximity matrix to aggregate the information of node attributes and links from neighboring nodes. Recently, some deep learning based methods \cite{kipf2016variational,pan2018adversarially,yang2016revisiting} try to learn the node representation by utilizing the neural network. HNE  \cite{chang2015heterogeneous} is a typical method that maps the nodes and attributes into the same latent space by utilizing the neural network for node representation learning. CSAN \cite{meng2020jointly} implements a jointly learning framework by embedding each node and attribute with means and variances of Gaussian distributions via variational auto-encoders. Although those methods achieve good performance in attribute network embedding, they do not take full advantage of attribute and structure information. Differently, in this paper, we design a jointly learning framework of the network in the hypersphere space which aims to capture the interaction between structure and attribute during the learning of the corresponding hyperspheres for anomaly detection task on attributed networks.

\subsection{Anomaly Detection on Attributed Networks}

Recently, anomaly detection on attributed networks has attracted lots of research interests \cite{fan2020anomalydae}, whose goal is to detect the anomalies by obtaining information from nodes attribute and network structure. Some of them conduct anomaly detection using only structure information in the community-level by comparing the current node with other reference nodes within the same community \cite{perozzi2016scalable} or measuring the quality of connected subgraphs \cite{perozzi2018discovering}. Some of them study the problem of feature-level anomalies detection through subspace selection of node feature \cite{sanchez2013subspaces, perozzi2014focused}. While some residual analysis based methods \cite{li2017radar, peng2018anomalous} and graph autoencoder based methods \cite{ding2019deep, li2019specae, fan2020anomalydae} consider both the context and feature information to discover anomalies by assuming that anomalies cannot be approximated from other reference nodes, they use residual estimation or network reconstruction to measure the abnormality of each node. However, most of those methods aim at minimizing the reconstruction errors for the whole network, whose loss functions are not designed directly for the abnormal node detection, moreover, those reconstruction based losses will also be impacted by the noisy node and have a potential problem of overfitting for both normal nodes and abnormal nodes. Differently, in this paper, we aim to design a new training objective towards measuring the abnormality of nodes from both the structure and attribute perspectives.

\begin{table}[bt]
\centering
\caption{Notations.}
\label{tab:notations}
\begin{threeparttable}
\begin{tabular}{l|l|l|l|}
\headrow
\thead{Notation} & \thead{Description}  &\thead{Notation} & \thead{Description} \\
\boldmath${\mathcal{G}}$    &  Attributed network.    &$r_{a}\in \boldsymbol{\mathcal{R}_{+}}$ & The radius of attribute hypersphere.    \\
\boldmath${\mathcal{V}}$  & A set of nodes in network.  &$r_{s}\in \boldsymbol{\mathcal{R}_{+}}$ & The radius of structure hypersphere. \\
\boldmath${\mathcal{A}}$ & A set of attributes in attributed network.  &$c_{a}\in \boldsymbol{\mathcal{R}}^{D}$ & The center of attribute hypersphere. \\
\boldmath${\mathcal{E}}$ & A set of edges in network.  &$c_{s}\in \boldsymbol{\mathcal{R}}^{D}$ & The center of structure hypersphere.  \\
$N$ & The number of nodes.   &\textup{$\boldsymbol{\mathrm{A}}\in \boldsymbol{\mathcal{R}}^{N \times N}$} & Adjacency matrix of a network. \\
$M$ & The dimension of attribute.  &\textup{$\boldsymbol{\mathrm{X}}\in \boldsymbol{\mathcal{R}}^{N \times M}$} & Attribute matrix of a network.\\
$D$ & The dimension of embedding.  &\textup{$\boldsymbol{\mathrm{Z}^{s}}\in \boldsymbol{\mathcal{R}}^{N \times D}$} & Node embedding in the structure space. \\
$\textbf{D}^{\mathcal{G}}$ &The diagonal degree matrix of network \boldmath${\mathcal{G}}$.  &\textup{$\boldsymbol{\mathrm{Z}^{a}}\in \boldsymbol{\mathcal{R}}^{M \times D}$} & Node embedding in the attribute space.\\
\hline  
\end{tabular}
\end{threeparttable}
\end{table}

\section{Notations and Problem Statement}
 In this section, we formally deﬁne the frequently-used notations and definitions used in this paper. The notations are summarized in Table \ref{tab:notations}.

\begin{myDef}
    Attributed network is denoted as $\boldsymbol{\mathcal{G}}=\{\boldsymbol{\mathcal{V}}, \boldsymbol{\mathcal{E}}, \textup{$\boldsymbol{\mathrm{X}}$} \}$ with $\emph{N}=|\boldsymbol{\mathcal{V}}|$ nodes and $|\boldsymbol{\mathcal{E}}|$ edges, $\textup{$\boldsymbol{\mathrm{X}}$}\subseteq{\boldsymbol{\mathcal{R}}^{N\times{D}}}$ denotes the attribute matrix of $\emph{N}$ nodes. The network structure can be represented as an adjacency matrix $\textup{$\boldsymbol{\mathrm{A}}$}\in{\boldsymbol{\mathcal{R}}^{\emph{N}\times\emph{N}}}$, where $\textup{$\boldsymbol{\mathrm{A}}$}_{i, j}=1$ if there is an edge between node $\emph{v}_i$ and $\emph{v}_j$, otherwise $\textup{$\boldsymbol{\mathrm{A}}$}_{i, j}=0$. 
\end{myDef}

\begin{mypro}
Given an attributed network $\boldsymbol{\mathcal{G}}=\{\boldsymbol{\mathcal{V}}, \boldsymbol{\mathcal{E}}, \textup{$\boldsymbol{\mathrm{X}}$} \}$, we aim at building a model to detect the nodes that differ significantly from the majority of the other nodes of $\boldsymbol{\mathcal{G}}$ in terms of both the attribute and structure patterns of the nodes. More formally, a score function $s: \boldsymbol{\mathcal{V}}_i \mapsto y_i \in \mathbb{R}$ is learned to classify the node $\boldsymbol{\mathcal{V}}_i$ based on the threshold $\lambda$:
\begin{equation}
\label{eq:problem}
\begin{split}
y_{i}=\left\{\begin{matrix}
1, & if \ s(\boldsymbol{\mathcal{V}}_i)\geq \lambda, \\ 
0, & otherwise.
\end{matrix}\right.
\end{split}
\end{equation}
where $y_{i}$ denotes the label of node $\boldsymbol{\mathcal{V}}_i$ , with 0 being the normal class and 1 the anomalous class.

\end{mypro}

\section{Method}

\begin{figure*}
\centering
\centerline{\includegraphics[width=5.5in]{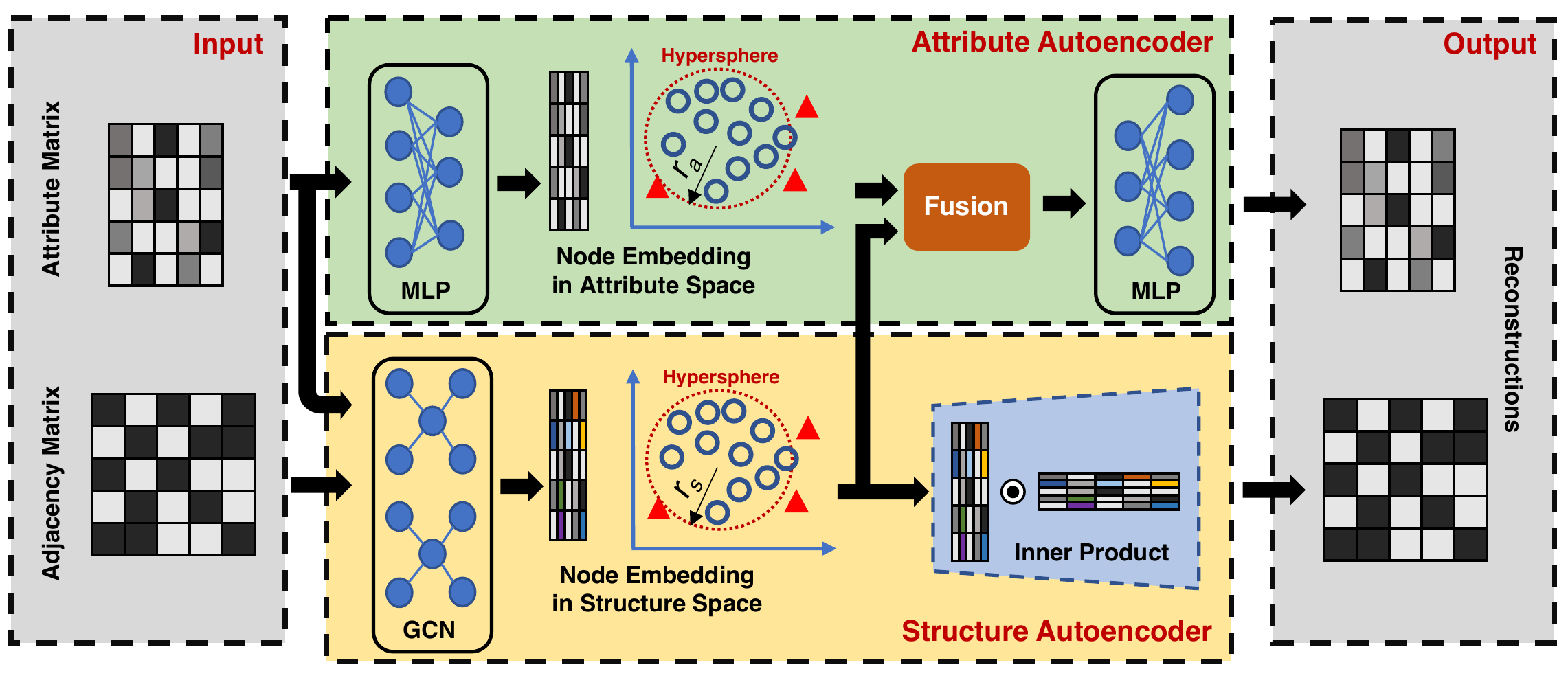}}
\caption{The framework of Dual-SVDAE.}
\label{fig:model}
\end{figure*}

In this section, we describe the proposed Dual-SVDAE in detail. The proposed Dual-SVDAE is an end-to-end framework for anomaly detection on attributed networks.  As shown in Fig.\ref{fig:model}, Dual-SVDAE consists of a structure autoencoder for structure hypersphere learning, an attribute autoencoder for attribute hypersphere learning. Moreover, to jointly learn the structure embedding and attribute embedding of nodes, we fuse the structure embedding and attribute embedding as the final input of the feature decoder to generate the node attribute. Finally, abnormal nodes are detected by measuring the distance of nodes to the learned center of each hypersphere in the latent structure space and attribute space respectively. 

\subsection{Structure Autoencoder}
Structure autoencoder takes both the adjacency matrix and attribute matrix as inputs to learn the structure hypersphere of normal nodes through a GCN encoder, and uses an inner product decoder for structure reconstruction. Specifically, in the structure encoder, we use graph convolutional layer (GCN) \cite{kipf2016semi} for structure feature extraction, where GCN convolution operation is defined as follows:
 \begin{equation}
\begin{split}
\label{eq:gcn}
&\textbf{GCN}(\textbf{X}, \textbf{A}|\boldsymbol{W})=\varphi((\textbf{D}^{\mathcal{G}})^{-\frac{1}{2}}\textbf{A}(\textbf{D}^{\mathcal{G}})^{-\frac{1}{2}}\textbf{X}\boldsymbol{W})
\end{split}
\end{equation}
where $\textbf{D}^{\mathcal{G}}$ denotes the diagonal degree matrix of network $\mathcal{G}$ with $(\textbf{D}^{\mathcal{G}})_{i,i}=\sum_{j=1}^{N}\textbf{A}_{i,j}$ being the degree of node $\mathcal{V}_i$. $\boldsymbol{W}$ is the trainable weight and $\varphi(\bullet)$ is activation function such as $\text{Relu}(x)=max(0, x)$ or $\text{Tanh}(x)=\frac{2}{1+e^{-2x}}-1$.

Then, a $L_s$-layers GCN encoder is employed to generate the node embedding $\textbf{Z}^{\textbf{s}}$ in the structure space: 
\begin{equation}
\label{eq:struc_encoder}
{\textbf{Z}^{\textbf{s}(l_s)}}=\textbf{GCN}(\textbf{Z}^{\textbf{s}(l_s-1)}, \textbf{A}|\boldsymbol{W}^{\textbf{s}(l_s)})
\end{equation}
where $\textbf{Z}^{\textbf{s}(l_s-1)}$ and $\textbf{Z}^{\textbf{s}(l_s)}$ are the input and output of $l_s$-th layer respectively, $l_s\in{\{1, 2, ..., L_s\}}$.  ${\textbf{Z}^{\textbf{s}(0)}}={\textbf{X}}$ is the initial input of the encoder and $\textbf{Z}^\textbf{s}=\textbf{Z}^{\textbf{s}(L_s)}$ is the final output of the encoder. $\boldsymbol{W}^{\textbf{s}(l_s)}$ is the learnable weights of the neural network. 

In the structure decoder, a simply inner product decoder is utilized to reconstruct the edges of the original network based on embedding similarity of each node pair. Specifically, structure decoder takes node embedding $\textbf{Z}^{\textbf{s}}$ as inputs and reconstruct the adjacency matrix $\hat{\textbf{A}}$ as follows:
\begin{equation}
\label{eq:struc_decoder}
\hat{\textbf{A}}=Sigmoid(\textbf{Z}^{\textbf{s}}(\textbf{Z}^{\textbf{s}})^{\rm{T}})
\end{equation}
where $Sigmoid(\bullet)$ is the sigmoid activation function, and $(\bullet)^{\rm{T}}$ is the transpose operation of a matrix. 

\subsection{Attribute Autoencoder}

Attribute autoencoder takes only the attribute matrix as input to learn the attribute hypersphere of normal nodes. Different from structure autoencoder, which uses GCN to learn the node representation by performing a message-passing mechanism on the network, in the attribute autoencoder, a simple multi-layer perceptron (MLP) is utilized in the encoder to learn a non-linear feature mapping of the node attribute information, which does not rely on the structure information, and uses the MLP based decoder for attribute reconstruction. Specifically, attribute encoder uses a $L^{e}_a$-layers MLP to generate the node embedding in the attribute space:
\begin{equation}
\label{eq:attr_encoder}
{\textbf{Z}^{\textbf{a}(l^e_a)}}=\sigma(\textbf{Z}^{\textbf{a}(l^e_a-1)}\boldsymbol{W}^{\textbf{\emph{x}}(l^e_a)}+\textbf{b}^{\textbf{\emph{x}}(l^e_a)})
\end{equation}
where $\textbf{Z}^{\textbf{a}(l^e_a-1)}$ and $\textbf{Z}^{\textbf{a}(l^e_a)}$ are the input and output of $(l^e_a)$-th layer respectively. $\boldsymbol{W}^{\textbf{\emph{x}}(l^e_a)}$ and $\textbf{b}^{\textbf{\emph{x}}(l^e_a)}$ are the trainable weights and bias of $(l^e_a)$-th layer. $l^e_a\in{\{1, 2, ..., L^e_a\}}$ and ${\textbf{Z}^{\textbf{a}(0)}}={\textbf{X}}$ is the initial input of the network. $\sigma(\bullet)$ denotes the activation function, such as ReLU. The output of attribute encoder is the final attribute embedding $\textbf{Z}^{\textbf{a}}=\textbf{Z}^{\textbf{a}(L^e_a)}$.

To achieve a joint learning for capturing the interaction between structure and attribute, following the attribute encoder, a feature fusion module is designed to fuse the learned node embedding $\textbf{Z}^{\textbf{s}}$ from structure space and $\textbf{Z}^{\textbf{a}}$ from attribute space into a fused embedding $\textbf{Z}^{\textbf{f}}$, which is taken as input by attribute decoder for attribute reconstruction. The fusion operation is defined as follows:
\begin{equation}
\label{eq:fusion_add}
\begin{split}
\textbf{Z}^{\textbf{f}}&=Fusion(\textbf{Z}^{\textbf{s}}, \textbf{Z}^{\textbf{a}})=\textbf{Z}^{\textbf{s}} \oplus \textbf{Z}^{\textbf{a}}
\end{split}
\end{equation}
where $\oplus$ indicates the element-wise plus operator of two matrices which add the corresponding elements in the same position of two matrices.

Based on the fused node embedding $\textbf{Z}^{\textbf{f}}$, in the attribute decoder, a $L^{d}_a$-layers MLP is utilized to reconstruct the attributes of the original network. Specifically, attribute decoder takes the fused node embedding $\textbf{Z}^{\textbf{f}}$ as inputs and reconstruct the attribute matrix $\hat{\textbf{X}}$ as follows:
\begin{equation}
\begin{split}
\label{eq:attr_decoder}
&\textbf{Z}^{\boldsymbol{\hat{\mathrm{X}}}(l^{d}_{a})}=\sigma(\textbf{Z}^{\boldsymbol{\hat{\mathrm{X}}}(l^{d}_{a}-1)}\textbf{W}^{\boldsymbol{\hat{\mathrm{X}}}(l^{d}_{a})}+\textbf{b}^{\boldsymbol{\hat{\mathrm{X}}}(l^{d}_{a})})
\end{split}
\end{equation}
where $\textbf{Z}^{\boldsymbol{\hat{\mathrm{X}}}(l^{d}_{a}-1)}$, $\textbf{Z}^{\boldsymbol{\hat{\mathrm{X}}}(l^{d}_{a})}$, $\textbf{W}^{\boldsymbol{\hat{\mathrm{X}}}(l^{d}_{a})}$ and $\textbf{b}^{\boldsymbol{\hat{\mathrm{X}}}(l^{d}_{a})}$ are the input, output, the trainable weight and bias matrix of ($l^{d}_{a}$)-th layer of decoder respectively, $l^{d}_{a}\in\{1,2,...,L^{d}_{a}\}$, and $\textbf{Z}^{\boldsymbol{\hat{\mathrm{X}}}(0)}=\textbf{Z}^{\textbf{f}}$ is the initial input of the decoder. Finally, the reconstruction $\boldsymbol{\hat{\mathrm{X}}}=\textbf{Z}^{\boldsymbol{\hat{\mathrm{X}}}(l^{d}_{a})}$ is obtained from the last layer of decoder.

\subsection{Hypersphere Learning}

Hypersphere learning aims at learning a compact hypersphere boundary to cover the normal data and detect outside data points far from the boundary. In this paper, we propose to learn two different hypersphere spaces for normal data from the structure and attribute perspectives respectively. The structure hypersphere parameterized by a center $c_s$ and a radius $r_s$, is inferred based on the learned node embedding $\textbf{Z}^{\textbf{s}}$ in the structure space, which are learned by Dual-SVDAE to minimize the average volumes of structure hypersphere as follows:
\begin{equation}
\label{eq:struc_hypersphere}
\underset{r_s}{\text{min}}\hspace{0.2cm} \mathcal{L}_s = r_s^2 + \frac{1}{\mu_{s} N}\sum_{i=1}^N \text{max}\{0, ||\textbf{Z}^{\textbf{s}}_i-c_s||^2-r_s^2\}
\end{equation}
where $\mu_{s}$ controls the trade-off between the volume of the structure hypersphere and violations of the boundary. Similarly, the attribute hypersphere parameterized by a center $c_a$ and a radius $r_a$, is inferred based on the learned node embedding $\textbf{Z}^{\textbf{a}}$ in the attribute space, which are learned by Dual-SVDAE to minimize the average volumes of attribute hypersphere as follows:
\begin{equation}
\label{eq:attr_hypersphere}
\underset{r_a}{\text{min}}\hspace{0.2cm} \mathcal{L}_a = r_a^2 + \frac{1}{\mu_a N}\sum_{i=1}^N \text{max}\{0, ||\textbf{Z}^{\textbf{a}}_i-c_a||^2-r_a^2\}
\end{equation}
where $\mu_{a}$ is also a hyperparameter that controls the trade-off between the volume of the attribute hypersphere and violations of the boundary.

The initialization and updating of $c_s$, $r_s$, $c_a$, and $r_a$ are described in the Algorithm \ref{alg:dual_svdae}. 

\subsection{Loss Function and Anomaly Score}
The training objective of Dual-SVDAE is to optimize the hypersphere and minimize the reconstruction error of both the network structure and attribute, which is defined as follows:
\begin{equation}
\label{eq:loss}
\mathcal{L}=\beta(\mathcal{L}_s+||\textbf{A}-\hat{\textbf{A}}||_{2}^{2})+(1-\beta)(\mathcal{L}_a+||\textbf{X}-\hat{\textbf{X}}||_{2}^{2})
\end{equation}
where $\beta$ controls the trade-off between the structure loss and attribute loss.
For a given test node $\boldsymbol{V}_i$, the anomaly score $s(\boldsymbol{V}_i)$ is defined as the distance of the point to the center of structure hypersphere and attribute hypersphere:
\begin{equation}
\label{eq:anomaly_socre}
s(\boldsymbol{V}_{i})=\beta([|| \textbf{Z}^{\textbf{a}}_i-c_a||^2-r_a^2]) +(1-\beta)(|| \textbf{Z}^{\textbf{s}}_i-c_s||^2-r_s^2])
\end{equation}
in this paper, the threshold $\lambda = 0$, and when  $s(\boldsymbol{V}_{i})>\lambda$, the node is classified as an anomaly. Otherwise, it is classified as a normal node.

\begin{algorithm}
\caption{Dual-SVDAE.} 
\label{alg:dual_svdae} 
\begin{algorithmic}[1] 
\REQUIRE ~~\\
Attributed network $\boldsymbol{\mathcal{G}}=\{\boldsymbol{\mathcal{V}}, \boldsymbol{\mathcal{E}}, \textbf{$\boldsymbol{\mathrm{X}}$}\}$; attributed encoder $f_a$, attributed decoder $f'_a$, structure encoder $f_s$, structure decoder $f'_s$.
\ENSURE ~~\\
$f_a$, $f_s$: the updated encoder; $r_a$, $r_s$: the updated hypersphere radius. \\
\STATE Initialize $r_s=0$, $r_a=0$, $c_s=\frac{1}{N}f_s(\boldsymbol{\mathcal{V}})$, $c_a=\frac{1}{N}f_a(\boldsymbol{\mathcal{V}})$.
\FOR{$epoch$ = 1 to $T$} 
\STATE Generate the node embedding $\textbf{Z}^{\textbf{s}}$ in the structure space via Eq. (\ref{eq:struc_encoder});
\label{step:struc_encoder}
\STATE Reconstruct the adjacency matrix $\textbf{A}$ based on the learned node embedding $\textbf{Z}^{\textbf{s}}$ via Eq. (\ref{eq:struc_decoder}).
\label{step:struc_decoder}
\STATE Generate the node embedding $\textbf{Z}^{\textbf{a}}$ in the attribute space via Eq. (\ref{eq:attr_encoder});
\STATE Fuse the learned node embeddings $\textbf{Z}^{\textbf{s}}$ from structure space and $\textbf{Z}^{\textbf{a}}$ from attribute space into a fused embedding $\textbf{Z}^{\textbf{f}}$ via Eq. (\ref{eq:fusion_add});
\label{step:fusion_add}
\STATE Reconstruct the attribute matrix $\textbf{X}$ based on the fused node embedding $\textbf{Z}^{\textbf{f}}$ via Eq. (\ref{eq:attr_decoder});
\label{step:attr_decoder}
\STATE Infer the structure hypersphere based on the learned node embedding $\textbf{Z}^{\textbf{s}}$ from the structure space via Eq. (\ref{eq:struc_hypersphere});
\label{step:struc_hypersphere}
\STATE Infer the attribute hypersphere based on the learned node embedding $\textbf{Z}^{\textbf{a}}$ from the attribute space via Eq. (\ref{eq:attr_hypersphere});
\label{step:attr_hypersphere}
\STATE Update the Dual-SVDAE with its stochastic gradient by Eq. (\ref{eq:loss}).
\label{step:loss}
\STATE Update $r_a$ as the distance of $(1-\mu_a) \times 100\%$ of all points to the center $c_a$ in the attribute space, and update $r_s$ as the distance of $(1-\mu_s) \times 100\%$ of all points to the center $c_s$ in the structure space.
\ENDFOR
\RETURN $f_a$, $f_s$, $r_a$, $r_s$.
\label{step:return}
\end{algorithmic}
\end{algorithm}

\section{EXPERIMENTS}

In this section, we firstly describe the experimental setups including datasets, baseline methods, parameter settings, and evaluation metrics. Then, we present the experimental results on the anomaly detection task in comparison with the state-of-the-art methods as well as ablation study and parameter analysis to evaluate the effectiveness of the proposed method.

\subsection{Datasets}
\label{subsec:Datasets}

In the experiment, following the previous study \cite{wang2021one}, three publicly used datasets Cora, Citeseer, and Pubmed are selected for evaluation. The statistics of datasets are shown in Table \ref{tab:datasets}. Details of the datasets are as follows: 

\begin{itemize}
\item \textbf{Cora}: Cora\footnote{{\url{https://relational.fit.cvut.cz/dataset/CORA}}} is a citation network, in which 2708 nodes $\boldsymbol{\mathcal{V}}$ are papers and 5429 edges $\boldsymbol{\mathcal{E}}$ are citation links among papers. The raw attribute $\boldsymbol{\mathcal{A}}$ of each node is represented as the 0/1 valued bag-of-words of the corresponding paper with dimension being equal to 1433. All papers are labeled by one of the following seven classes: Case Based (CB), Genetic Algorithms (GA), Neural Networks (NN), Probabilistic Methods (PM), Reinforcement Learning (RL), Rule Learning (RuL), and Theory (TH) respectively.

\item \textbf{Citeseer}: Citeseer\footnote{{\url{https://linqs-data.soe.ucsc.edu/public/lbc/citeseer.tgz}}} is a citation network, in which 3327 nodes $\boldsymbol{\mathcal{V}}$ are scientific publications and 4732 edges $\boldsymbol{\mathcal{E}}$ are citation links among those publications. The raw attribute $\boldsymbol{\mathcal{A}}$ of each node is described as the 0/1 valued bag-of-words of the corresponding publications with dimension being equal to 3703. All publications are labeled by one of the following six classes: Agents, Artificial Intelligence (AI), Database (DB), Information Retrieval (IR), Machine Learning (ML), and Human Computer Interaction (HCI).

\item \textbf{Pubmed}: Pubmed\footnote{{\url{https://linqs-data.soe.ucsc.edu/public/Pubmed-Diabetes.tgz}}} is a scientific publications network pertaining to diabetes, in which 19717 nodes $\boldsymbol{\mathcal{V}}$ are publications and 44338 edges $\boldsymbol{\mathcal{E}}$ are citation links among publications. The raw attribute $\boldsymbol{\mathcal{A}}$ of each node is represented as the TF/IDF weighted word vector of the corresponding publications with dimension being equal to 500. All papers are labeled by one of the following three classes: Diabetes Mellitus Experimental (Exp.), Diabetes Mellitus Type 1 (Type1), and Diabetes Mellitus Type 2 (Type2).
\end{itemize}

Following the standard unsupervised anomaly detection settings \cite{ruff2018deep, wang2021one}, we select one of the classes as the normal class, and the samples from the remaining classes are used as anomalies. The training set includes only the normal class, while in the validation set and the test set, half of the nodes are normal and the other half are randomly sampled from the anomalous class. For normal class nodes, we split 60\% for training and split 15\% and 25\% for validation and test respectively.

\begin{table}
\centering
\caption{Statistics of the used real-world datasets.}
\label{tab:datasets}
\begin{threeparttable}
\begin{tabular}{l|l|l|l|l}
\headrow
\thead{Dataset} & \thead{$\boldsymbol{\mathcal{V}}$}  & \thead{$\boldsymbol{\mathcal{E}}$}  & \thead{$\boldsymbol{\mathcal{A}}$} & \thead{Class}\\
Cora           &2, 708  &5, 429    &1, 433 &7\\
Citeseer       &3, 327  &4, 732    &3, 703 &6\\
Pubmed         &19, 717 &44, 338   &500   &3\\
\hline  
\end{tabular}
\end{threeparttable}
\end{table}

\subsection{Baselines}

We compare three types of baselines to evaluate the effectiveness of the proposed method. The first type is the traditional one-class classification based methods including OC-SVM \cite{scholkopf2001estimating} and Deep-SVDD \cite{ruff2018deep}. The second type is the graph autoencoder based methods including GAE \cite{kipf2016variational} and Dominant \cite{ding2019deep}. The third type is the latest one-class graph neural network based model OC-GNN \cite{wang2021one}. Details of those baselines are as follows:

\begin{itemize}
    \item \textbf{OC-SVM (RAW)} \cite{scholkopf2001estimating}: OC-SVM (Raw) is a classical anomaly detection method that conducts OC-SVM \cite{scholkopf2001estimating} on the raw node attribute for anomaly detection. 
    \item \textbf{OC-SVM (DW)} \cite{scholkopf2001estimating}: Different from OC-SVM (Raw), OC-SVM (DW) employs the graph embedding algorithm DeepWalk \cite{perozzi2014deepwalk} to represent node in the embedding space based only on the structure information, and then conducts one-class classification on the generated node embedding for anomaly detection.
    \item \textbf{Deep-SVDD (Attr)} \cite{ruff2018deep}: Deep-SVDD is a neural network based one-class classification method inspired by the classic kernel-based Support Vector Data Description (SVDD), in which a data-enclosing hypersphere of normal data points is learned for anomaly detection. In this paper, Deep-SVDD (Attr) learns the hypersphere of normal node by using node attribute as the sample feature for model training.
    \item \textbf{Deep-SVDD (Stru)} \cite{ruff2018deep}: Different from Deep-SVDD (Attr), Deep-SVDD (Stru) learns the hypersphere of normal data by using the structure information among nodes as the sample feature for model training.
    \item \textbf{GAE} \cite{kipf2016variational}: GAE is a GCN \cite{kipf2016semi} based autoencoder, which leverages both structure and attribute information for node embedding, and uses node embedding to reconstruct the original network topology structure. In the experiment, we select the reconstruction error as anomaly score for anomaly detection.
    \item \textbf{Dominant} \cite{ding2019deep} is a dual autoencoder based method that detects anomalies by ranking both the structure and attribute reconstruction errors learned by GCN.
    \item \textbf{OC-GNN} \cite{wang2021one} is an one-class classification method based on GNNs such as GCN \cite{kipf2016semi} (termed as OC-GCN) or GraphSage \cite{hamilton2017inductive} (termed OC-SAGE) for anomaly detection on attributed networks. OC-GNN combines the powerful representation ability of graph neural networks along with the classical one-class objective to conduct anomaly detection on graph structure data.
\end{itemize}

\subsection{Experimental Design and Metrics}

In the experiment, we implement Dual-SVDAE based on Pytorch\footnote{\url{https://pytorch.org/}} and Deep Graph Library \footnote{\url{https://www.dgl.ai/}}, and train it with 5000 iterations for Cora and 2000 iterations for Citeseer and Pubmed datasets respectively. Adam algorithm is utilized for optimization with learning rate as 0.002. The embedding dimension is set as 32 for all datasets. The hyperparameter $[\mu_a, \mu_s, \beta]$ set as [0.2, 0.9, 0.2], [0.4, 0.6, 0.4] and [0.4, 0.9, 0.2] for Cora, Citeseer and Pubmed respectively. For the baseline methods, we use the publicly available implementations from the original papers and set the parameters by grid search. 

We evaluate all methods under 5 different seeds and the average performance and the mean-variance are reported as the final results. Similar to the previous studies \cite{ding2019deep, wang2021one}, in this paper, we use the widely used \textbf{AUC} score (the \textbf{A}rea \textbf{U}nder a receiver operating characteristic \textbf{C}urve) and \textbf{AP} score (\textbf{A}verage \textbf{P}recision) to measure the anomaly detection performance of different models.  All the experiments are conducted on the Ubuntu 16.04 Server with 128 GB memory, Intel(R) Xeon(R) CPU E5-2640 (2.4 GHz), and GeForce RTX 2080 Ti Graphics Card.

\begin{table}[bt]
\centering
\caption{Anomaly Detection Performance. All values are percentages (\%) and averaged with StdDevs over 5 seeds. The best results are marked in \textbf{bold}.}
\label{tab:anomaly_detection}
\begin{threeparttable}
\resizebox{\textwidth}{!}{
\begin{tabular}{cV{1}c|cV{1}c|cV{1}c|c}
\hlineB{1}
\headrow
\textbf{Dataset} &\multicolumn{2}{cV{1}}{{\textbf{Cora}}} &\multicolumn{2}{cV{1}}{{\textbf{Citeseer}}}   &\multicolumn{2}{c}{{\textbf{Pubmed}}} \\
\textbf{Metrics} & \textbf{AUC} & \textbf{AP} & \textbf{AUC} & \textbf{AP} & \textbf{AUC} & \textbf{AP}  \\
\hlineB{1}
OCSVM (Raw) \cite{scholkopf2001estimating}      &60.94$\pm$5.37 &57.52$\pm$4.46  &55.39$\pm$3.81 &54.82$\pm$2.77   &54.50$\pm$9.62 &53.72$\pm$7.26  \\
OCSVM (DW) \cite{scholkopf2001estimating}  &66.86$\pm$12.92 &64.63$\pm$11.94  &50.17$\pm$ 12.58 &51.94$\pm$ 8.08  &52.38$\pm$8.43   &52.37$\pm$6.06  \\
Deep-SVDD (Attr) \cite{ruff2018deep}  &63.80$\pm$3.30 &62.20$\pm$3.51  &67.08$\pm$7.43 &66.60$\pm$7.91 &70.80$\pm$9.43 &71.07$\pm$10.55 \\
Deep-SVDD (Stru) \cite{ruff2018deep}  &70.23$\pm$4.52 &69.20$\pm$5.31  &73.42$\pm$10.13 &73.69$\pm$9.74 &71.38$\pm$9.80 &71.66$\pm$11.03 \\
GAE \cite{kipf2016semi} &56.43$\pm$7.45 &59.13$\pm$7.76 &52.27$\pm$7.53  &52.50$\pm$5.62   &53.79$\pm$6.37  &53.16$\pm$4.32 \\
Dominant \cite{ding2019deep} &87.19$\pm$3.70 &78.45$\pm$6.24 &80.07$\pm$7.71 &72.32$\pm$0.15 &65.49$\pm$2.22 &61.76$\pm$2.43\\
OC-GCN \cite{wang2021one} &68.19$\pm$11.19 &66.76$\pm$9.15 &75.29$\pm$6.55 &73.29$\pm$6.35 &66.65$\pm$5.37 &68.05$\pm$6.22 \\
OC-SAGE \cite{wang2021one} &90.99$\pm$2.60 &89.37$\pm$3.12 &84.13$\pm$6.67 &81.52$\pm$5.53 &74.60	$\pm$6.17  &75.11$\pm$6.83 \\
\textbf{Dual-SVDAE }      &\textbf{95.09$\pm$2.50} &\textbf{95.05$\pm$2.66}  &\textbf{86.54$\pm$6.72}	&\textbf{86.20$\pm$7.08} &\textbf{91.32$\pm$2.37}  &\textbf{92.39$\pm$2.55} \\

\hlineB{1}  
\end{tabular}}
\end{threeparttable}
\end{table}

\subsection{Experimental Results}

\subsubsection{Performance Analysis}

In this section, we demonstrate the effectiveness of the proposed method by presenting the results of our model on the anomaly detection task, and provide a comparison with the state-of-the-art methods. The results of all compared methods on Cora, Citeseer, and Pubmed datasets are provided in Table \ref{tab:anomaly_detection}.

The experimental results show that the proposed Dual-SVDAE significantly outperforms all baselines across all the datasets. For example, compared with the best baseline OC-SAGE, Dual-SVDAE increases the AUC score and AP score by 5.10\% and 5.68\% on Cora dataset, 2.41\% and 4.68\% on Citeseer dataset, and 16.72\% and 17.28\% on Pubmed dataset. Among the traditional methods of OCSVM(Raw), and Deep-SVDD (Attr), they consider only the attribute information of nodes, while ignoring the network structure information of the attribute network, and therefore results in poor performance. Different from those traditional methods, GAE and Dominant leverage both structure and attribute information for node embedding, however, the reconstruction error used in those graph autoencoder based models cannot effectively measure the abnormality. For those one-class graph based methods like OC-GCN and OC-SAGE, they achieve better performance than other baselines, the main reason lies in the powerful graph structure feature extraction such as the simple neighbor aggregation strategy used in OC-SAGE and the hypersphere learning mechanism which is designed directly for anomaly detection. However, different from OC-GCN and OC-SAGE, our proposed Dual-SVDAE is capable of measuring the anomalies from both the structure and attribute perspectives through a dual-hypersphere learning mechanism.

\begin{figure*}[bt]
\centering
\subfigure[Cora]{\includegraphics[height=1.3in]{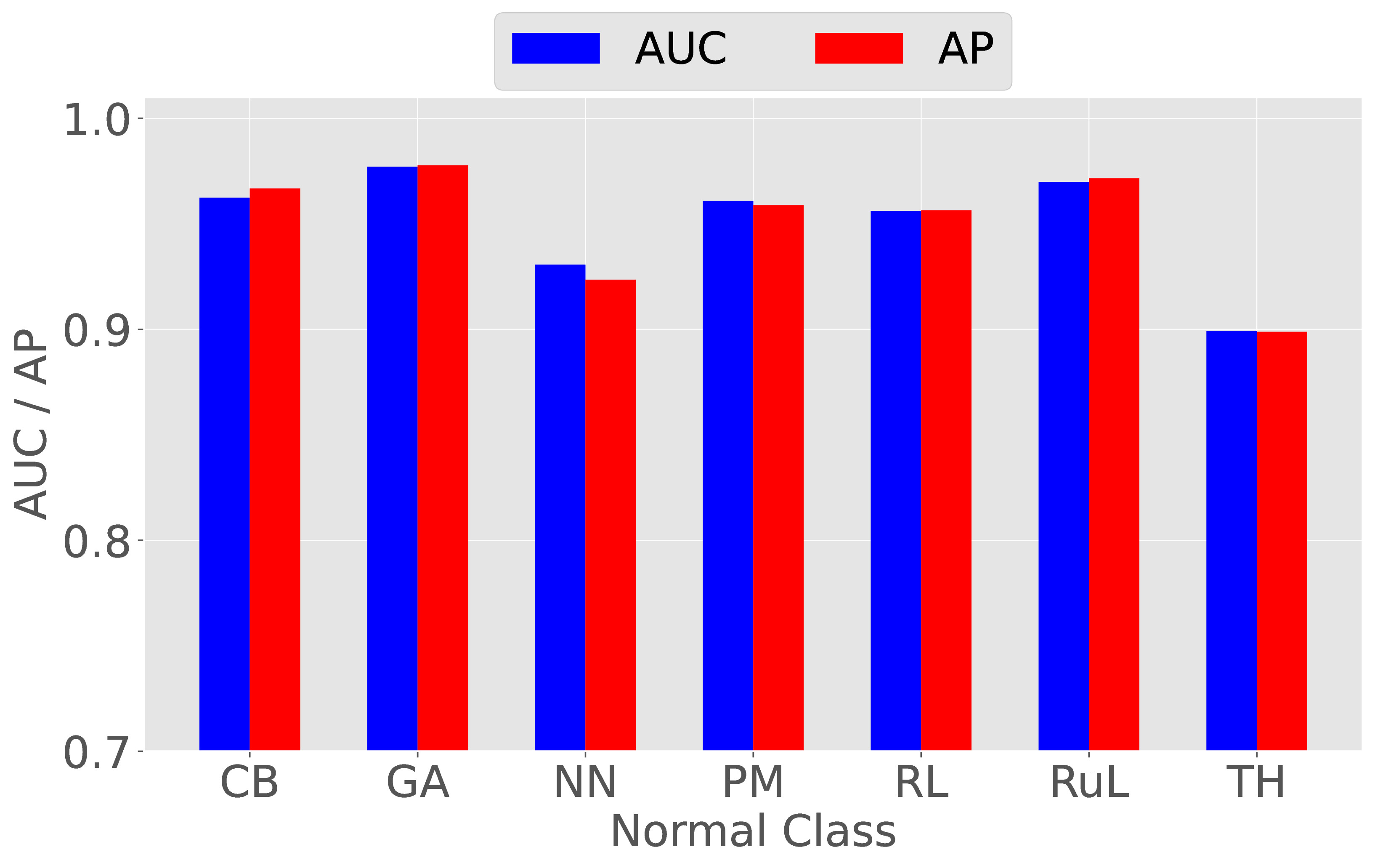}}
\subfigure[Citeseer]{\includegraphics[height=1.3in]{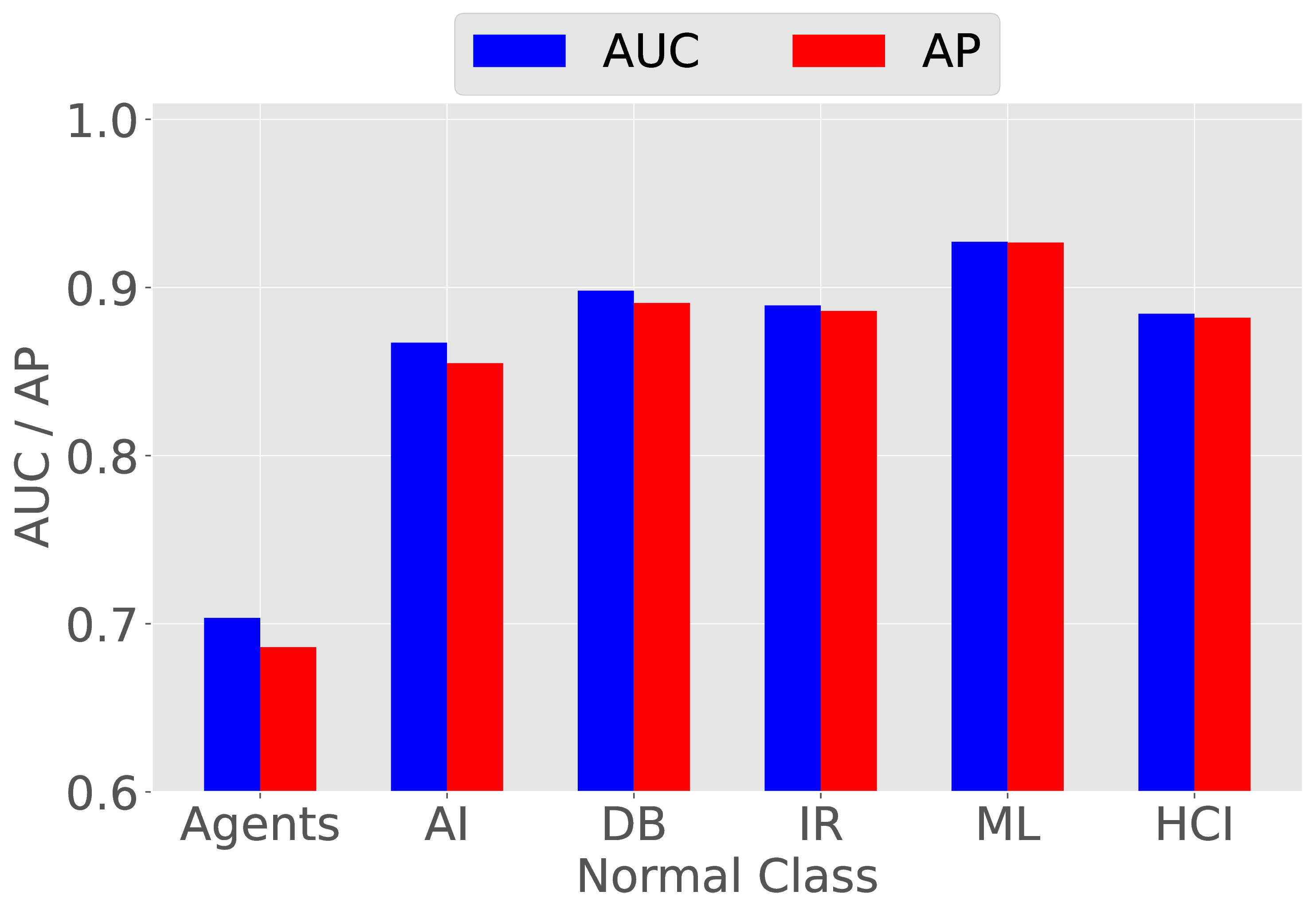}}
\subfigure[Pubmed]{\includegraphics[height=1.3in]{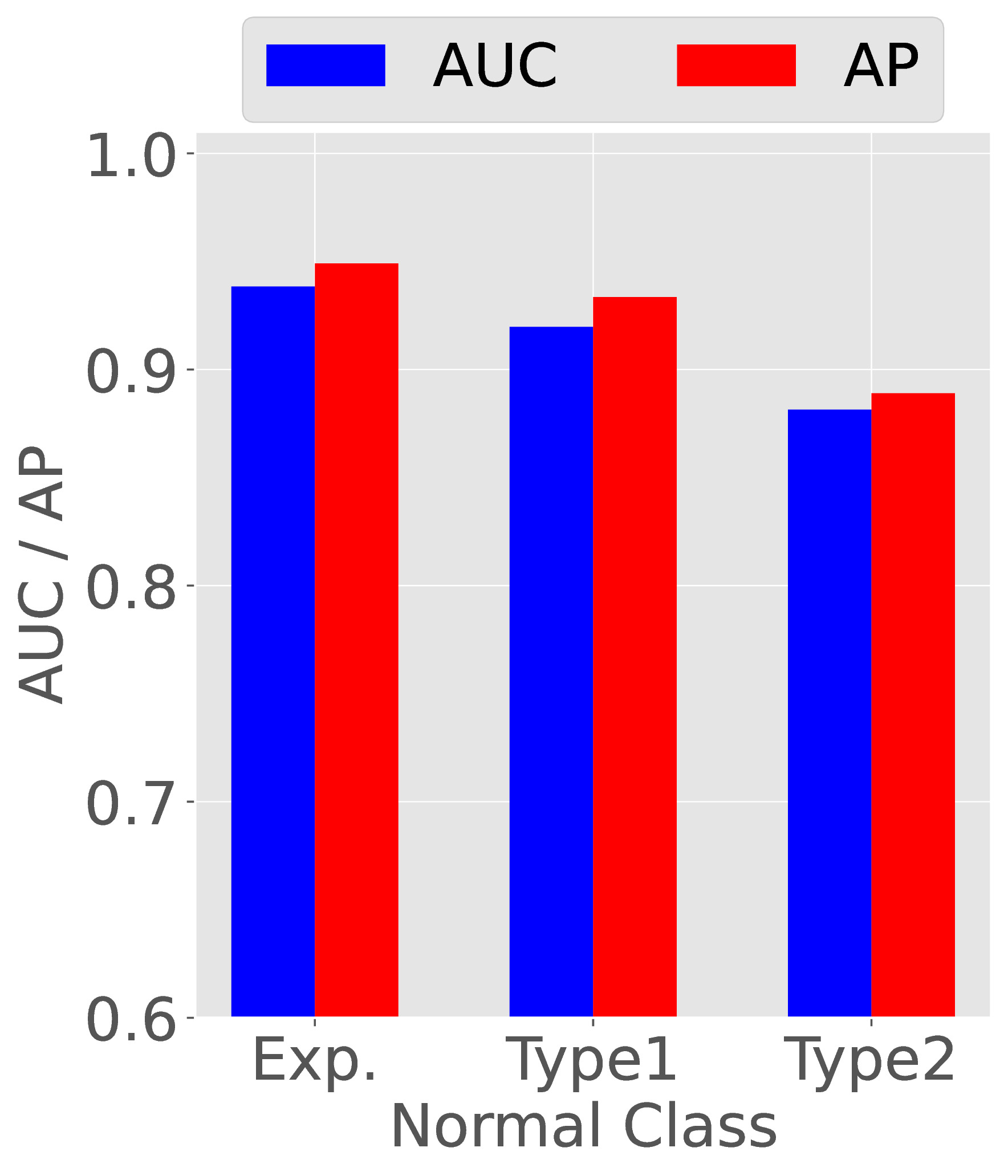}}
\caption{Evaluation performance of Dual-SVDAE by taking each class of nodes as normal class on Cora, Citeseer, and Pubmed datasets.}
\label{fig:normal_class}
\end{figure*}

Fig.\ref{fig:normal_class} shows the results of Dual-SVDAE by taking each class of nodes as a normal class on three datasets. Overall, we find that Dual-SVDAE performs stable for each class on three datasets, which demonstrates the robustness and effectiveness of Dual-SVADE. Interestingly, on the Citeseer dataset, the detection performance is slightly worse on the "Agents" class compared with other classes, one possible reason behind this might be that there are only 7.35\% Agents-type nodes in the original network, which cannot provide enough training samples for model learning.

\subsubsection{Parameter Sensitive Analysis}

\begin{figure*}
\centering
\subfigure[Result of AUC score.]{\includegraphics[width=2.7in]{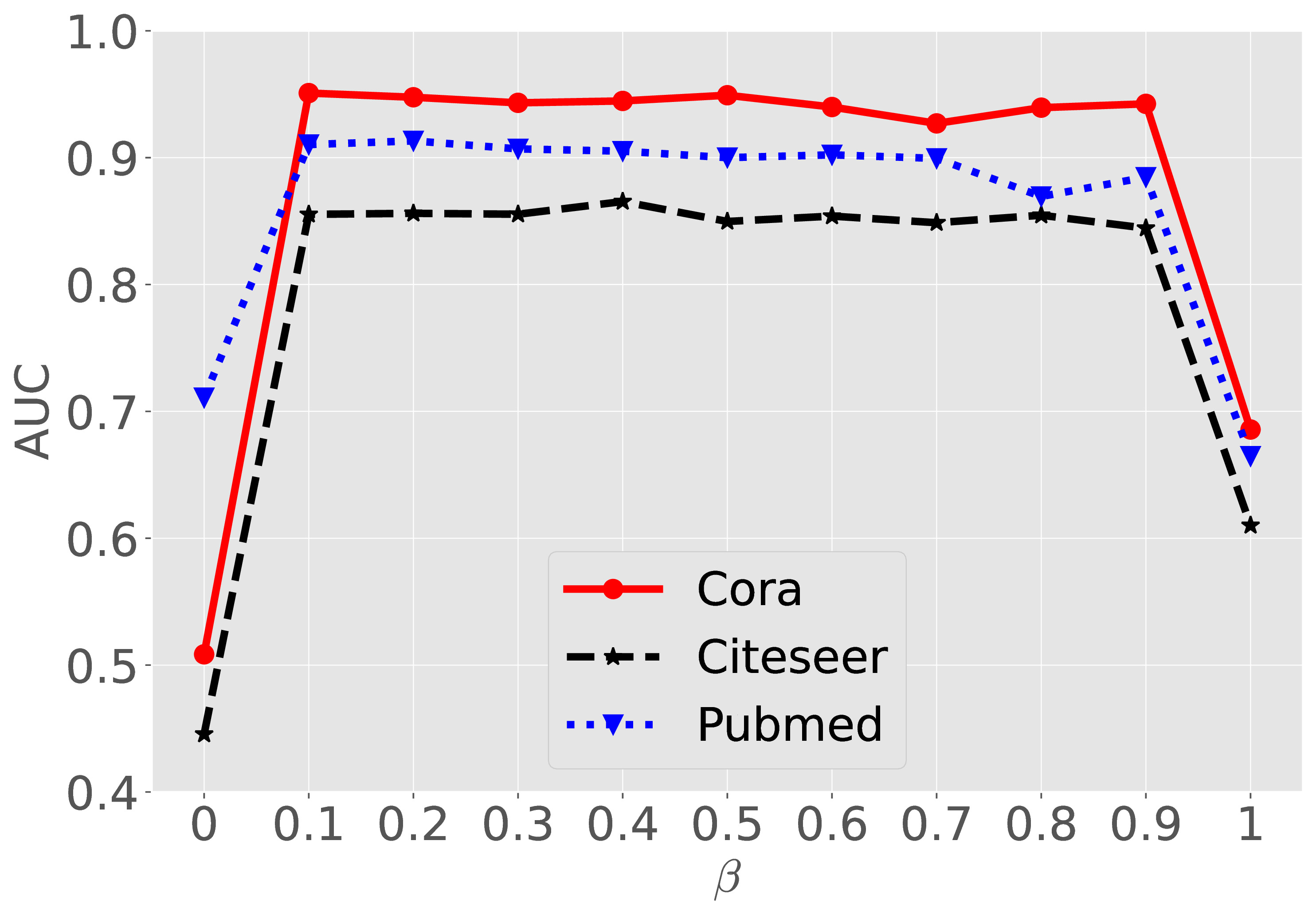}}
  \subfigure[Result of AP score]{\includegraphics[width=2.7in]{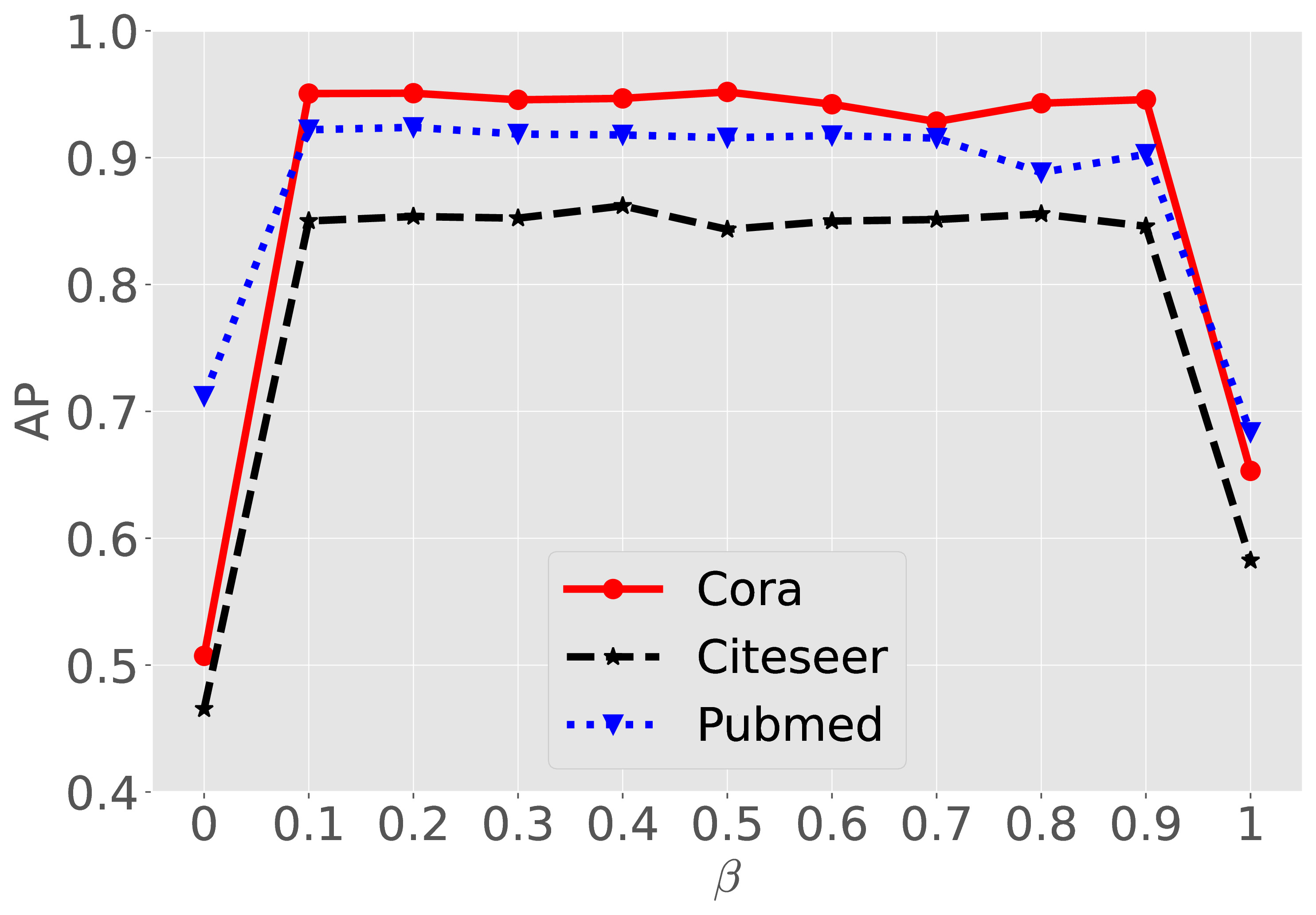}}
\caption{Impact of different  trade-off weights $\beta$ between the structure anomaly and attribute anomaly measurements on Cora, Citeseer, and Pubmed datasets. }
\label{fig:sensitivity_beta}
\end{figure*}

In this section, we investigate the impact of different trade-off weights $\beta$ between structure and attribute anomaly measure on three datasets. Fig.\ref{fig:sensitivity_beta} shows the impact of different trade-off weights $\beta$ between structure and attribute anomaly measure on Cora, Citeseer, and Pubmed datasets. Firstly, we find that Dual-SVDAE performs stable when $0<\beta<1$, which demonstrates its advantage of parameters insensitivity. Moreover, by only measuring the attribute abnormality $(\beta=0)$ or the structure abnormality$(\beta=1)$, Dual-SVDAE cannot achieve an effective detection for different types of networks, because when $\beta=0$ or $\beta=1$, the structure autoencoder or attribute autoencoder does not participate in the model training, therefore, the anomalies are measured only under the structure or attribute perspective, which results in poor detection performance.

\subsubsection{Ablation Study}

In this section, we investigate the impact of node attribute, network structure, and hypersphere learning on Dual-SVDAE for anomaly detection. Specifically, we study the contribution of different modules in Dual-SVDAE independently. The details of ablation settings are as follows:

\begin{itemize}
    \item \textbf{wo-$OC$}: we remove the hypersphere learning module of Dual-SVDAE and use two reconstruction losses of network structure and node attribute respectively, to supervise the training of the model. Finally, we use reconstruction error as anomaly score for anomaly detection.
    \item \textbf{wo-$AE_{s}$}: we remove the structure autoencoder module and use only the attribute autoencoder for training, and conduct anomaly detection based on the hypersphere in the attribute space.
    \item \textbf{wo-$AE_{a}$}: we remove the attribute autoencoder module and use only the structure autoencoder for training, and conduct anomaly detection based on the hypersphere in the structure space.
    \item \textbf{wo-$De_{a}$}: we remove the attribute decoder module during learning the hypersphere in the attribute space.
    \item \textbf{wo-$De_{s}$}: we remove the structure decoder module during learning the hypersphere in the structure space.
    \item \textbf{wo-$De_{both}$}: we remove both the structure decoder and attribute decoder during learning the hypersphere in the latent space.
\end{itemize}

Tabel \ref{tab:ablation_study} shows the results of ablation study on three datasets. We find that Dual-SDVAE achieves the best results. The poor results of wo-$OC$, wo-$AE_s$, and wo-$AE_a$ demonstrate the effectiveness of the proposed hypersphere learning mechanism. Moreover, when we only use the structure feature or attribute feature, the model performs worse, one possible reason behind this might be that considering only the attribute information or structure information will destroy the integrity of the information for the attributed network. Therefore, considering both the structure information and attribute information is necessary for anomaly detection on attributed network. The models without decoder including wo-$De_{a}$, wo-$De_{s}$ and wo-$De_{both}$ perform better than wo-$OC$, wo-$AE_s$ and wo-$AE_a$, but they perform worse than Dual-SVDAE, which demonstrate the effectiveness of the proposed joint learning mechanism and that the reconstruction regularization on the node embedding is important for high-quality embedding generation. Based on the above analysis, all the modules of Dual-SVDAE have positive effects for anomaly detection on attributed networks.

\begin{table}
\centering
\caption{Ablation study of Dual-SVDAE. All values are percentages (\%) and averaged with StdDevs over 5 seeds.}
\label{tab:ablation_study}
\begin{threeparttable}
\resizebox{\textwidth}{!}{
\begin{tabular}{cV{1}c|cV{1}c|cV{1}c|c}
\hlineB{1}
\headrow
\textbf{Dataset} &\multicolumn{2}{cV{1}}{{\textbf{Cora}}} &\multicolumn{2}{cV{1}}{{\textbf{Citeseer}}}   &\multicolumn{2}{c}{{\textbf{Pubmed}}} \\
\textbf{Metrics}  &\textbf{AUC} & \textbf{AP}  & \textbf{AUC} &\textbf{AP} & \textbf{AUC} & \textbf{AP} \\
\hlineB{1}
Dual-SVDAE  &95.09$\pm$2.50 &95.05$\pm$2.66 &86.54$\pm$6.72	&86.20$\pm$7.08 &91.32$\pm$2.37 &92.39$\pm$2.55  \\
wo-$OC$  &55.94$\pm$3.80 &53.08$\pm$2.37  &56.85$\pm$2.46   &59.51$\pm$5.67 &54.01$\pm$2.01 &55.10$\pm$2.31  \\
wo-$AE_{s}$ &50.85$\pm$1.57 &50.73$\pm$2.08 &44.56$\pm$3.69  &46.55$\pm$2.37  &71.08$\pm$7.71 &71.19$\pm$9.28 \\
wo-$AE_{a}$ &68.57$\pm$9.23 &65.31$\pm$9.67 &61.01$\pm$6.20  &58.25$\pm$6.05  &66.47$\pm$5.86 &68.35$\pm$7.02 \\
wo-$De_{a}$ &93.45$\pm$3.16 &93.95$\pm$3.09 &84.18$\pm$6.48  &84.18$\pm$6.36  &82.29$\pm$8.61 &84.08$\pm$9.57\\
wo-$De_{s}$ &76.16$\pm$3.45 &73.42$\pm$4.58 &64.69$\pm$3.46 &61.40$\pm$3.16  &65.26$\pm$4.59 &66.41$\pm$5.33 \\
wo-$De_{both}$ &77.15$\pm$4.38 &74.01$\pm$5.03 &65.91$\pm$4.20  &61.93$\pm$4.32  &66.87$\pm$3.32 &66.97$\pm$5.15 \\
\hlineB{1}

\end{tabular}}
\end{threeparttable}
\end{table}

\begin{figure*}
\centering
  \subfigure[OC-SVM (DW)]{\includegraphics[width=5.5in]{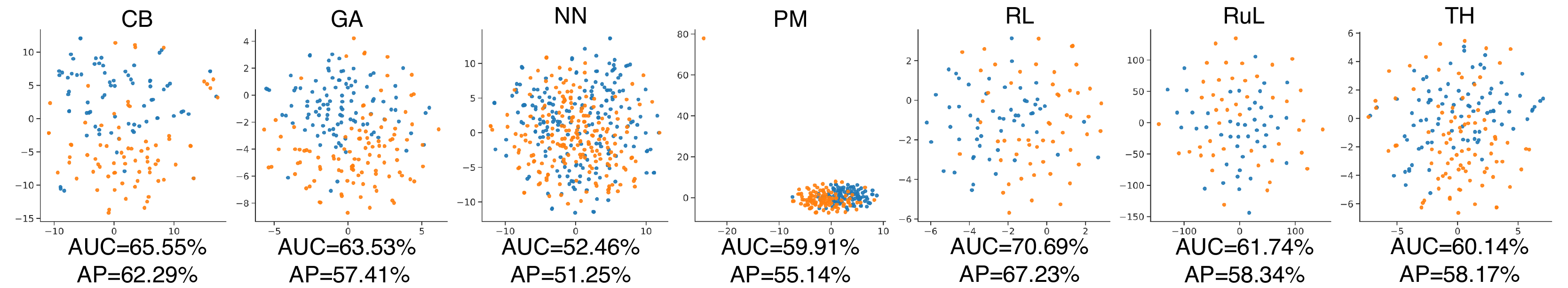}}
  \subfigure[OC-GCN]{\includegraphics[width=5.5in]{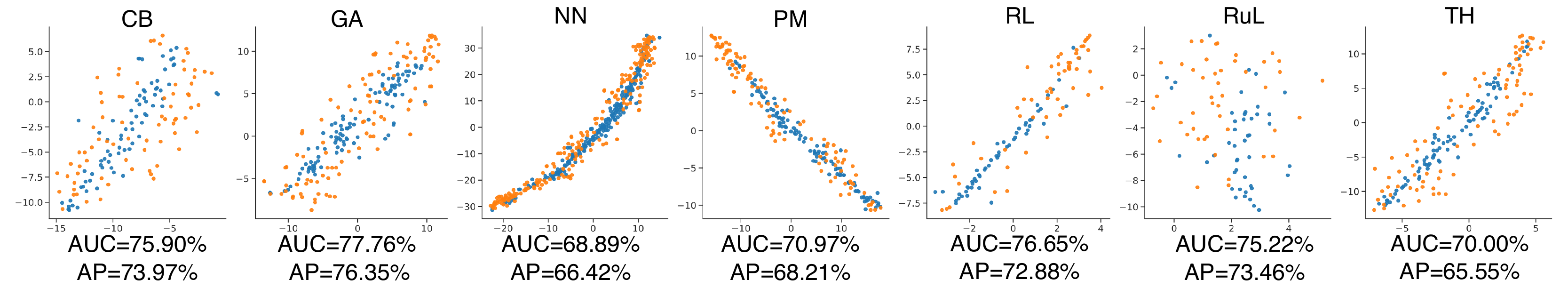}}
  \subfigure[Dual-SVDAE]{\includegraphics[width=5.5in]{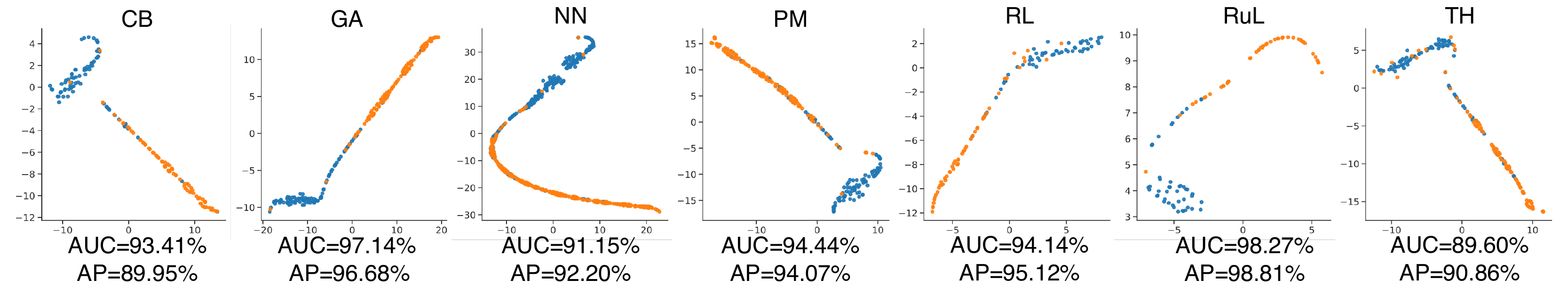}}
  \caption{T-SNE visualization of different node embedding by taking each class of nodes as normal class on Cora dataset. (\BL{\textbf{Blue}} points indicate the normal nodes and \textcolor{orange}{\textbf{orange}} points indicate the abnormal nodes respectively.)}
\label{vis}
\end{figure*}

\begin{figure}
\centering
\subfigure{\includegraphics[width=5.5in]{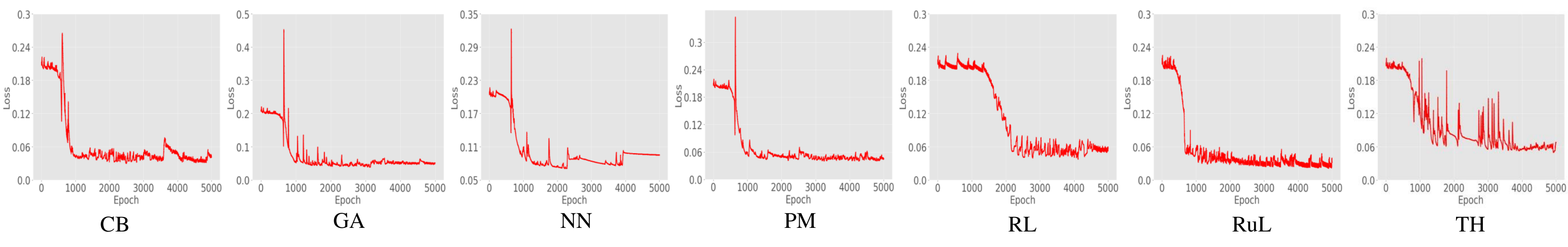}}
\caption{Learning curve of Dual-SVDAE on Cora dataset.}
\label{loss_change}
\end{figure}

\subsubsection{Visualization and Convergence Analysis}

In this section, we make a comparison of the visualization of node embedding for different methods in Fig.\ref{vis} and conduct convergence analysis of Dual-SVDAE by plotting the learning curve of training loss in Fig. \ref{loss_change}. 

For visualization of node embedding, specifically, we take the low-dimensional embeddings of test nodes learned by OC-SVM (DW), OC-GCN, and Dual-SVDAE respectively, as the input of the t-SNE tool \cite{van2008visualizing}, which maps the learned node embedding into a two-dimensional space. Blue points indicate the normal nodes and orange points indicate the abnormal nodes respectively. As we can see, comparing with OC-SVM(DW) and OC-GCN, the normal nodes are clustered and separated better from those abnormal nodes by Dual-SVDAE, which demonstrates that the dual-hypersphere learning mechanism facilitates better node embedding to capture both the attribute and structure patterns among nodes. The visualized results also explain why Dual-SVDAE achieves the better performance of anomaly detection on attributed networks.

For convergence analysis, we plot the training loss of Dual-SVDAE on the Cora dataset. Specifically, Fig.\ref{loss_change} shows the learning curve of Dual-SVDAE by taking each class of nodes as a normal class on the Cora dataset. Overall, we find that Dual-SVDAE performs stable for each class on all classes of Cora, which demonstrates the convergence of the model's feature learning and shows the stability of Dual-SVDAE.

\section{Conclusion}

In this paper, we study the problem of anomaly detection on attributed networks. To cope with the problem, we proposed an end-to-end model Deep Dual Support Vector Data Description (Dual-SVDAE)  for anomaly detection on attributed networks. Dual-SVDAE learns the compact hypersphere boundary of normal nodes' latent space from both the structure and attribute perspectives, and abnormal nodes can be detected by measuring the distance of nodes to the learned center of each hypersphere in the latent structure space and attribute space respectively. Extensive experiments on multiple real-world attributed networks demonstrate quantitatively as well as qualitatively the effectiveness of the proposed method.

\bibliography{refs}



\end{document}